%% file: 0_main.tex
\documentclass{article}

\usepackage{xcolor}
\usepackage{colortbl}
\usepackage{microtype}
\usepackage{graphicx}
\usepackage{subcaption}
\usepackage{colortbl}
\usepackage{booktabs}
\usepackage{booktabs} % for professional tables
\usepackage{hyperref}
\usepackage{xurl}
\definecolor{lightblue}{RGB}{173, 216, 230}
\usepackage{cuted}
\usepackage{caption}
\usepackage{graphicx}
\usepackage{enumitem}
\usepackage[preprint]{icml2026}

\usepackage{amsmath}

\usepackage{amssymb}
\usepackage{mathtools}
\usepackage{amsthm}

\usepackage[capitalize,noabbrev]{cleveref}

\theoremstyle{plain}

\theoremstyle{definition}

\theoremstyle{remark}

\usepackage[textsize=tiny]{todonotes}
\usepackage{multirow}
\usepackage{booktabs}
\usepackage{bbm}
\usepackage{multicol}
\usepackage{float}
\usepackage{cuted}
\icmltitlerunning{CORAL: Transferrable and Calibration-Aware Inference-Time Steering}

\begin{document}

\twocolumn[
  \icmltitle{Correctness-Optimized Residual Activation Lens (CORAL): \\
  Transferrable and Calibration-Aware Inference-Time Steering}



  \icmlsetsymbol{equal}{*}

  \begin{icmlauthorlist}
    \icmlauthor{Miranda Muqing Miao}{yyy}
    \icmlauthor{Young-Min Cho}{yyy}
    \icmlauthor{Lyle Ungar}{yyy}
  \end{icmlauthorlist}

  \icmlaffiliation{yyy}{Department of Computer and Information Science,  University of Pennsylvania, Philadelphia, USA}

  \icmlcorrespondingauthor{Miranda Muqing Miao}{miaom@seas.upenn.edu}

  \icmlkeywords{Mechanistic Interpretability, Probe Steering, LLM}

  \vskip 0.3in
]



\printAffiliationsAndNotice{}  % no special notice (required even if empty)

% \input{sections/01_abstract}
\input{sections/01_abstract}

\input{sections/02_introduction}
\input{sections/03_methodology}
\input{sections/04_coral_results}
\input{sections/05_confidence_results}

\input{sections/06_conclusion}

\input{sections/07_limitations}

% \section*{Impact Statement}

% Authors are \textbf{required} to include a statement of the potential broader
% impact of their work, including its ethical aspects and future societal
% consequences. This statement should be in an unnumbered section at the end of
% the paper (co-located with Acknowledgements -- the two may appear in either
% order, but both must be before References), and does not count toward the paper
% page limit. In many cases, where the ethical impacts and expected societal
% implications are those that are well established when advancing the field of
% Machine Learning, substantial discussion is not required, and a simple
% statement such as the following will suffice:

% ``This paper presents work whose goal is to advance the field of Machine
% Learning. There are many potential societal consequences of our work, none
% which we feel must be specifically highlighted here.''

% The above statement can be used verbatim in such cases, but we encourage
% authors to think about whether there is content which does warrant further
% discussion, as this statement will be apparent if the paper is later flagged
% for ethics review.

\bibliography{bib}
\bibliographystyle{icml2026}

%%%%%%%%%%%%%%%%%%%%%%%%%%%%%%%%%%%%%%%%%%%%%%%%%%%%%%%%%%%%%%%%%%%%%%%%%%%%%%%
%%%%%%%%%%%%%%%%%%%%%%%%%%%%%%%%%%%%%%%%%%%%%%%%%%%%%%%%%%%%%%%%%%%%%%%%%%%%%%%
% APPENDIX
%%%%%%%%%%%%%%%%%%%%%%%%%%%%%%%%%%%%%%%%%%%%%%%%%%%%%%%%%%%%%%%%%%%%%%%%%%%%%%%
%%%%%%%%%%%%%%%%%%%%%%%%%%%%%%%%%%%%%%%%%%%%%%%%%%%%%%%%%%%%%%%%%%%%%%%%%%%%%%%
\newpage
\appendix
\onecolumn

\input{sections/appendix}

% TBU
%%%%%%%%%%%%%%%%%%%%%%%%%%%%%%%%%%%%%%%%%%%%%%%%%%%%%%%%%%%%%%%%%%%%%%%%%%%%%%%
%%%%%%%%%%%%%%%%%%%%%%%%%%%%%%%%%%%%%%%%%%%%%%%%%%%%%%%%%%%%%%%%%%%%%%%%%%%%%%%

\end{document}

%% file: sections/01_abstract.tex
\begin{abstract}
Large language models (LLMs) exhibit persistent miscalibration, especially after instruction tuning and preference alignment. Modified training objectives can improve calibration, but retraining is expensive. Inference-time steering offers a lightweight alternative, yet most existing methods optimize proxies for correctness rather than correctness itself. We introduce CORAL (Correctness-Optimized Residual Activation Lens), a regularized inference-time steering method that captures distributed correctness signals from model internal activations using weight-decay MLP probes. We evaluate CORAL across three 7B-parameter models and find that it consistently improves accuracy by 10\% and expected calibration error (ECE) by 50\% on average. We additionally demonstrate that these gains transfer without retraining to the complete published test sets of four held-out benchmarks (ARC-Challenge, HellaSwag, Math-MC, OpenBookQA), averaging 14\% accuracy improvements and 49\% ECE improvements. Our results support the hypothesis that distributed information in model internals can be extracted using regularized probes when individual neurons are insufficient. CORAL thus provides a compute-efficient, transferable, and calibration-aware approach to improve MCQA performance during inference.
\end{abstract}

%% file: sections/02_introduction.tex
\section{Introduction}

Inference-time steering has emerged as a promising approach for modifying LLM behavior without retraining overhead. Most steering methods target behavioral goals such as honesty, refusal, or stylistic attributes \citep{rimsky2024steering}. In this paper, we focus on a more direct objective: steering for factuality while simultaneously improving calibration. This dual goal is important because miscalibration persists across models and post-training pipelines, and common alignment methods like Reinforcement Learning with Human Feedback (RLHF) and Direct Policy Optimization (DPO) worsen calibration even as they improve other metrics \citep{jiang2021know, leng2025taming, ouyang2022training, rafailov2023direct}.

% Steering methods broadly divide into two paradigms. Sparse approaches search for discrete, localized features, often using Sparse Autoencoders (SAEs) to identify individual neurons that causally influence behavior \citep{bricken2023monosemanticity, templeton2024scaling}. Other steering approaches instead operate over distributed representations, treating the relevant signal as spread across many dimensions \cite{zou2023representation}. Recent mechanistic work suggests that while some behaviors localize to interpretable circuits, many others, including those related to confidence and correctness, are distributed and cannot be fully captured by individual features \citep{anthropic2024circuitsupdatesjuly, lindsey2025biology}. We focus on the distributed setting and show that regularized probes can extract correctness signals that sparse methods miss.

Steering methods divide into two paradigms. Sparse approaches search for discrete, localized features, often using Sparse Autoencoders (SAEs) to identify individual neurons that causally influence behavior \citep{bricken2023monosemanticity, templeton2024scaling}. Distributed approaches instead treat the relevant signal as spread across many dimensions \citep{zou2023representation}.

Recent mechanistic work suggests that confidence and correctness signals fall into the distributed category. While some behaviors localize to interpretable circuits, many others cannot be fully captured by individual features \citep{anthropic2024circuitsupdatesjuly, lindsey2025biology}. We focus on the distributed setting and show that regularized probes extract correctness signals that sparse methods miss.

Within distributed steering for factuality and calibration, three recent methods are most relevant: Inference-Time Intervention (ITI) \citep{li2024inference}, SteerConf \citep{zhou2025steerconf}, and Calibrating LLM Confidence by Probing Perturbed Representation Stability (CCPS) \citep{khanmohammadi-etal-2025-calibrating}. ITI performs distributed steering on attention head activations to improve truthfulness, but its effect on factual correctness remains underexplored. SteerConf elicits verbalized confidence through prompting and uses expressed confidence as a proxy for calibration. CCPS measures representation stability under adversarial perturbations to predict correctness, achieving calibration improvements and occasional accuracy gains. These methods share two limitations: they optimize proxies for correctness rather than correctness itself, and they do not leverage internal activations to directly steer model outputs. Furthermore, none of them demonstrates transferability to out-of-distribution benchmarks. 

In this paper, we address these gaps with CORAL, a regularized inference-time steering method that captures distributed correctness signals from model internal activations using weight-decay MLP probes. We train a regularized probe that predicts correctness directly from frozen residual stream activations, then use these predictions to steer model behavior at inference time. By jointly optimizing for the accuracy and calibration components of the Brier score, CORAL improves both metrics and transfers its impact successfully to unseen benchmarks without any retraining.

Our contributions are as follows:
\begin{enumerate}
    \item We introduce CORAL, a lightweight weight-decay MLP probe that consistently improves both accuracy and calibration across three 7B-parameter models and multiple MCQA benchmarks, using only 8.4k--10k training questions.
    \item We demonstrate that CORAL transfers without retraining to the complete published test sets of four held-out benchmarks (ARC-Challenge, HellaSwag, Math-MC, OpenBookQA), averaging 14\% accuracy improvements and 49\% ECE improvements. This transfer indicates that the probe captures a general-purpose correctness subspace rather than task-specific patterns.
    \item We provide evidence that correctness signals are distributed across many neurons. SAE-based ablations show negligible individual neural effects, while regularized probes successfully aggregate these distributed signals into effective steering vectors.
\end{enumerate}

\section{Related Works}

\textbf{Probing and internal representations.} Linear probes have long been used to decode task-relevant variables from intermediate representations \citep{alain2017understanding}. Research on transformer feed-forward layers shows that residual streams contain rich, interpretable information that accumulates across layers \citep{geva2021transformer, geva2022transformer}. Recent studies demonstrate that LLM activations can predict answer correctness from question-only signals, though this work focuses on prediction rather than intervention \citep{cencerrado2025noanswer}.

\textbf{Inference-time intervention (ITI).} ITI methods modify activations during forward passes to shift model behavior, and attention-head interventions have improved truthfulness by targeting a small set of heads \citep{li2024inference}. Contrastive activation addition instead trains steering vectors from positive and negative examples to control behavioral attributes \citep{rimsky2024steering}. These methods demonstrate that activation edits can causally influence outputs, but they typically target softer behavioral goals rather than factuality directly.

\textbf{Sparse Autoencoders and mechanistic interpretability.} SAEs decompose activations into sparse, approximately monosemantic features \citep{bricken2023monosemanticity, cunningham2024sparse}. Practitioners use SAE features for steering or ablation because they appear more interpretable than raw neurons \citep{templeton2024scaling, gao2024scaling}. However, circuit tracing work reveals that many model behaviors arise from distributed computation across many features rather than isolated units \citep{conmy2023towards, ameisen2025circuit, lindsey2025biology}. Our SAE ablation experiments confirm this finding for correctness signals: individual features produce negligible causal effects.

\textbf{Calibration in LLMs.} Neural networks, including LLMs, often exhibit miscalibration where predicted confidence fails to match empirical accuracy \citep{guo2017calibration, jiang2021know}. Post-hoc methods like temperature scaling and isotonic regression adjust output probabilities but cannot improve accuracy \citep{guo2017calibration, 10.1145/775047.775151}. Recent work identifies entropy neurons that regulate confidence without affecting task performance \citep{stolfo2024confidence}, and shows that calibration information evolves across layers \citep{joshi2025calibration}. CORAL differs from these approaches by using internal activations to jointly improve accuracy and calibration through direct residual prediction. 

\begin{figure}[t]  
    \centering
    \includegraphics[width=0.5\textwidth]{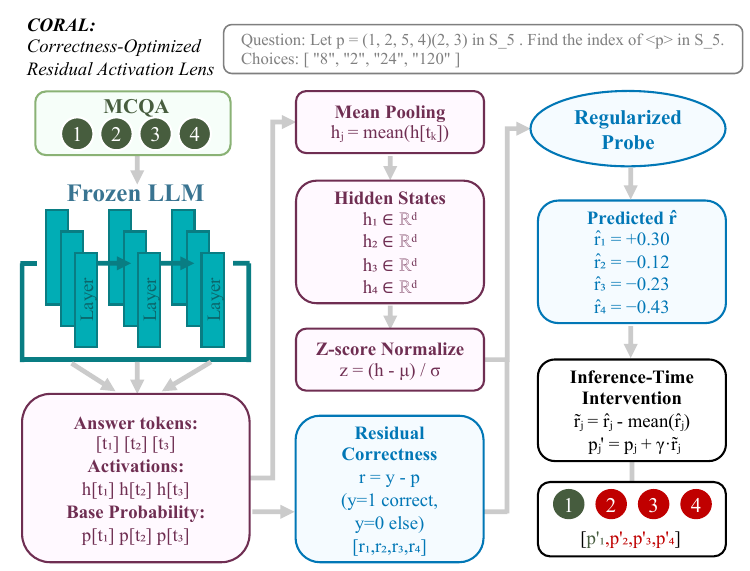}  
    \vspace{-8pt}
    \captionsetup{skip=2pt}
    \caption{Overview of CORAL. Given an MCQA question, a frozen LLM produces per-option hidden states and base probabilities. Hidden states are mean-pooled over answer tokens, z-score normalized, and passed through a trained probe to predict residual correctness. At inference, predicted residuals are centered and used to steer base probabilities toward better calibration.}
    \label{fig:coral_fig}  
    \vspace{-21pt}
\end{figure}

%% file: sections/03_methodology.tex
\section{Methodology}

Having established that existing methods optimize proxies rather than correctness directly, we now describe CORAL's methodology. Probing literature shows that intermediate representations contain rich task-relevant information beyond what surfaces in final predictions. We exploit this insight by training lightweight probes on frozen activations to predict residual correctness: the gap between a model's assigned probability and the ideal target. This formulation directly optimizes the Brier score and enables inference-time steering that improves both accuracy and calibration without modifying model weights.

We first formalize residual correctness (\ref{sec:problem_formulation}), then describe CORAL's architecture (\ref{sec:coral_architecture}), and finally detail our experimental setup (\ref{sec:experimental_setup}).

\subsection{Problem Formulation: Residual Correctness}
\label{sec:problem_formulation}

Consider a multiple-choice question answering task where a model assigns probability $p_j$ to each option $j \in \{1, 2, \ldots, n\}$. We define residual correctness $r_j$ as the gap between the ideal target distribution and the model's prediction:
\begin{equation}
r_j = \begin{cases}
1 - p_j & \text{if option } j \text{ is correct} \\
-p_j & \text{if option } j \text{ is incorrect}
\end{cases}
\end{equation}
Positive residuals indicate underconfidence on correct answers; negative residuals indicate overconfidence on incorrect answers. A perfectly calibrated model has $r_j = 0$ for all options.

This formulation directly targets the Brier score, a proper scoring rule. The Brier score decomposes as $\sum_j (p_j - y_j)^2 = \sum_j r_j^2$, where $y_j \in \{0,1\}$ indicates correctness. Training a probe to predict $\hat{r}_j$ and applying the correction $p'_j = p_j + \hat{r}_j$ minimizes expected squared error between predictions and ground truth. CORAL thus optimizes a principled calibration objective rather than fitting an empirical pattern.

\subsection{CORAL Architecture}
\label{sec:coral_architecture}

CORAL consists of three components: activation extraction, a regularized MLP probe, and inference-time steering.

\subsubsection{Activation Extraction}
\label{sec:activation_extraction}

For each question with $n$ answer options, we perform $n$ forward passes, concatenating the question prompt with each candidate answer. We record hidden states $\mathbf{h}_{j,t}^{(l)} \in \mathbb{R}^{d}$ at layer $l$ while the model processes answer tokens, then mean-pool over token positions:
\begin{equation}
\mathbf{h}_j^{(l)} = \frac{1}{T_j} \sum_{t=T_{\text{prompt}}+1}^{T_{\text{prompt}}+T_j} \mathbf{h}_{j,t}^{(l)}
\end{equation}
 The resulting dataset contains $N \times 4$ activation vectors, where $N$ is the number of questions. Each vector has dimensionality $d_{\text{model}}$ (e.g., 4096 for 7B parameter models). We pair each activation with its corresponding residual correctness label $r_j$, computed from the model's softmax probabilities and ground truth. This structure preserves the grouped nature of the data, where each question contributes exactly $n$ samples corresponding to its answer options. In the end, we apply z-score normalization using training set statistics before passing activations to the probe.

\subsubsection{Weight-Decay MLP Probe}
\label{sec:mlp_probe}

We train a four-hidden-layer MLP (dimensions 1024, 512, 256, 128) with ReLU activations, dropout (p=0.2), and tanh output to bound predictions to $[-1, 1]$. The loss combines mean squared error with an output penalty:
\begin{equation}
\mathcal{L} = \frac{1}{N} \sum_{i=1}^{N} (r_i - \hat{r}_i)^2 + \lambda_{\text{out}} \cdot \frac{1}{N} \sum_{i=1}^{N} \hat{r}_i^2
\end{equation}
We optimize using AdamW with weight decay, selecting hyperparameters via grid search over validation $R^2$. Training requires under 5 hours on a single GPU.

\subsubsection{Inference-Time Steering}
\label{sec:steering}

At inference, we extract activations for each answer option, compute probe predictions $\hat{r}_j$, center them to ensure zero-sum, and apply additive corrections:
\begin{equation}
p'_j = \frac{\max(p_j + \gamma \cdot \tilde{r}_j, 0)}{\sum_{j'=1}^{n} \max(p_{j'} + \gamma \cdot \tilde{r}_{j'}, 0)}
\end{equation}
where $\tilde{r}_j = \hat{r}_j - \frac{1}{n}\sum_{j'} \hat{r}_{j'}$ and $\gamma$ controls steering strength. This shifts probability mass toward options the probe identifies as correct while maintaining a valid distribution.

\subsection{Experimental Setup}
\label{sec:experimental_setup}

\subsubsection{Training Datasets}

We train two independent probes on two sets of datasets to demonstrate the generalization and robustness of CORAL. 

 \paragraph{Construct Training Dataset.} We first construct a custom training dataset by aggregating questions from two reasoning benchmarks. Specifically, we evenly sample 5,000 questions each from CommonsenseQA and RACE \textbf{training} questions, yielding 10,000 total training questions \cite{talmor-etal-2019-commonsenseqa, lai-etal-2017-race}. We partition this construct dataset into 80\% train / 20\% validation at the question level. We use GroupKFold to ensure all options from the same question remain in the same split, preventing information leakage across partitions. We then test on those benchmarks' corresponding and complete \textbf{test} questions. We call the probe trained on this Construct Training Dataset \textbf{Probe 1.}
 
\paragraph{MMLU Split Dataset.} We also train a probe based on Massive Multitask Language Understanding (MMLU), which is a benchmark that tests the models' knowledge and reasoning on a broad range of topics, using multiple choice questions \cite{hendrycks2021measuring}. We partition the 14k test questions from MMLU into training (8,400 questions, 60\%), validation (2,800 questions, 20\%), and evaluation (2,800 questions, 20\%) splits.
This is because MMLU provides no in-distribution training split. The dev set contains only 1,531 questions, which is insufficient for training probes over 4,096-dimensional activations. The 90k auxiliary train dataset of MMLU contains a vast number of question included in other benchmarks. Training on the 90k auxiliary dataset would therefore contaminate our transfer experiments. We call the probe trained on this MMLU Split Dataset \textbf{Probe 2.} 

We partition the 14k test questions from MMLU into training (8,400 questions, 60\%), validation (2,800 questions, 20\%), and evaluation (2,800 questions, 20\%) splits.

\paragraph{Prompting.} We collect activations strictly following few-shot prompting template from EleutherAI \texttt{lm-eval} harness templates \cite{eval-harness}. For each question, we extract mean-pooled hidden states over answer tokens at each layer.

\paragraph{Validation.} We solely use the validation data sets to determine the optimal steering layer and steering strength $\gamma$. 

\subsubsection{Transfer Benchmarks}
We evaluate generalization on on the complete published test sets of four held-out benchmarks: HellaSwag, OpenBookQA, Math-MC, and ARC-Challenge \citep{li-etal-2025-hellaswag, banerjee-etal-2019-careful, amini-etal-2019-mathqa, clark2018thinksolvedquestionanswering}. These benchmarks share no questions with training data and evaluate CORAL's helpfulness topics spanning commonsense NLP completion, multi-step question answering, math reasoning, and abstraction reasoning. 

\subsubsection{Evaluation Metrics}

We report accuracy and four calibration metrics: Expected Calibration Error (ECE), class-wise ECE (cwECE), Brier score, and negative log-likelihood (NLL). ECE measures alignment between confidence and accuracy across binned predictions. The Brier score captures both accuracy and calibration as mean squared error against ground truth. Lower values indicate better performance for all calibration metrics. 

% We define all metrics formally in Appendix~\ref{sec:metrics}.

\paragraph{Accuracy.}
For each question, we select the answer option with the highest probability and compare it to the ground truth label:
\begin{equation}
\text{Accuracy} = \frac{1}{N} \sum_{i=1}^{N} \mathbf{1}\left[\arg\max_j p_{i,j} = y_i\right]
\end{equation}
where $N$ is the number of questions, $p_{i,j}$ is the probability assigned to option $j$ for question $i$, and $y_i$ is the correct answer.

\paragraph{Expected Calibration Error (ECE).}
ECE measures the alignment between predicted confidence and actual accuracy. We partition predictions into $B$ equally-spaced bins based on the maximum probability (confidence) and compute the weighted average of the gap between accuracy and confidence within each bin:
\begin{equation}
\text{ECE} = \sum_{b=1}^{B} \frac{|S_b|}{N} \left| \text{acc}(S_b) - \text{conf}(S_b) \right|
\end{equation}
where $S_b$ is the set of samples whose confidence falls in bin $b$, $\text{acc}(S_b)$ is the accuracy of samples in bin $b$, and $\text{conf}(S_b)$ is the average confidence in bin $b$. Lower ECE indicates better calibration. A perfectly calibrated model should have ECE = 0. For all statistics reported, the chosen bin count $B$ = 25 unless otherwise stated. 

\paragraph{Class-wise Expected Calibration Error (cwECE).}
While ECE only considers the top predicted class, cwECE evaluates calibration across all answer options. For each class $c \in \{1, \ldots, n\}$, we compute a per-class ECE by binning the predicted probabilities $p_{i,c}$ and comparing them to the binary correctness indicator $\mathbf{1}[y_i = c]$:
\begin{equation}
\text{ECE}_c = \sum_{b=1}^{B} \frac{|S_{b,c}|}{N} \left| \text{acc}_c(S_{b,c}) - \text{conf}_c(S_{b,c}) \right|
\end{equation}
where $S_{b,c}$ contains samples whose probability for class $c$ falls in bin $b$, $\text{acc}_c(S_{b,c})$ is the fraction of samples in the bin where class $c$ is correct, and $\text{conf}_c(S_{b,c})$ is the average probability for class $c$ in the bin. The overall cwECE is the average across all classes:
\begin{equation}
\text{cwECE} = \frac{1}{n} \sum_{c=1}^{n} \text{ECE}_c
\end{equation}

For all statistics reported, the chosen bin count $B$ = 25 unless otherwise stated. 

\textbf{Brier Score.} The Brier score measures the mean squared error between predicted probabilities and ground truth labels. For each question $i$ with predicted distribution $\mathbf{p}_i = (p_{i,1}, \ldots, p_{i,n})$ and correct answer $y_i$, the Brier score is:
\begin{equation}
    \text{Brier} = \frac{1}{N} \sum_{i=1}^{N} \sum_{j=1}^{n} \left( p_{i,j} - \mathbbm{1}[j = y_i] \right)^2
\end{equation}
where $\mathbbm{1}[j = y_i]$ equals 1 if option $j$ is correct and 0 otherwise. Lower Brier scores indicate better calibrated and more accurate predictions. Unlike ECE, the Brier score is a proper scoring rule that jointly rewards accuracy and calibration.

\textbf{Negative Log-Likelihood (NLL).} NLL measures the negative log-probability assigned to the correct answer:
\begin{equation}
    \text{NLL} = -\frac{1}{N} \sum_{i=1}^{N} \log p_{i, y_i}
\end{equation}
where $p_{i, y_i}$ is the probability assigned to the correct option for question $i$. Lower NLL indicates higher confidence on correct answers. NLL heavily penalizes confident incorrect predictions, making it sensitive to calibration failures.

\subsubsection{Baselines}
We compare CORAL against established steering and calibration methods. These include (1) Inference Time Intervention (ITI)~\citep{li2024inference}, which performs activation steering at attention heads rather than residual stream positions; (2) SteerConf~\citep{zhou2025steerconf}, which targets confidence elicitation as a proxy for correctness; (3) Contrastive Confidence Probing and Steering (CCPS)~\citep{khanmohammadi-etal-2025-calibrating}, which uses contrastive learning on correct versus incorrect activations. We also report Few-Shot Prompting results using the Eval Harness framework~\citep{eval-harness}. Detailed descriptions of baseline implementations are provided in Appendix~\ref{sec:baselines}.

%% file: sections/04_coral_results.tex
\section{CORAL Results}
\input{mmlu_table}
\input{transfer_table}

In this section, we present CORAL's impact on in-distribution test questions, a layer analysis, and CORAL's success on out-of-distribution transfer benchmarks.

\label{sec:results}
\subsection{CORAL Steering Results}

CORAL steering produces consistent gains in both accuracy and calibration across all three models. Table~\ref{tab:steering_results} reports single-layer results for each model, with optimal layers selected via validation only.

We sweep steering strength $\gamma \in \{0.25, 0.5, \ldots, 2.75, 3.0\}$ on the validation set. The parameter $\gamma$ controls the magnitude of activation shifts along the learned correctness direction. We find that $\gamma = 1$ produces the strongest and most consistent results, suggesting that unit-magnitude shifts align well with the natural scale of residual stream activations.

CORAL yields leading performance across accuracy and most calibration metrics. On RACE, CORAL outperforms the second-best method by 17.39 percentage points in accuracy and achieves a 64\% relative improvement in ECE using Deepseek-LLM-7B-Chat. On CommonsenseQA and MMLU, CORAL also consistently achieves the lowest Brier scores across all three model families while maintaining meaningfully competitive accuracy.

% Regarding architecture, we find that MLPs with one hidden layer slightly under-perform linear probes by 0.5\%-0.75\%, while deeper architectures yield no further gains. This indicates that extracting residual correctness signals requires only modest nonlinearity, consistent with the hypothesis that correctness information is encoded in relatively accessible directions within the activation space.

Probe architecture has minimal impact on performance. MLPs with one to three hidden layer underperform linear probes by 0.5\%--0.75\%, while deeper architectures yield no further gains. This pattern indicates that extracting residual correctness signals requires only modest nonlinearity.

% The accessibility of correctness information explains this result. Correctness signals appear encoded in relatively accessible directions within the activation space, allowing simple architectures to capture them effectively.

% Multi-layer probes provide no additional benefit. We trained probes on concatenated activations from the top 3--5 layers to capture signals potentially distributed across processing stages. This approach produced no meaningful improvement over single-layer steering, suggesting that correctness-relevant information concentrates sufficiently at individual layers. Aggregating across layers introduces noise without providing additional signal.

We also investigated multi-layer CORAL, where probes are trained on concatenated activations from the top 3-5 layers. The motivation was to capture correctness signals that might be distributed across the model's later processing stages. However, this approach produced no meaningful improvement over single-layer steering. This suggests that correctness-relevant information is sufficiently concentrated at individual layers, and aggregating across layers does not providing additional meaningful signals.

\subsection{Layer Analysis}

% Layer-wise steering performance for DeepSeek-7B-Chat on MMLU. Left: accuracy across layers 0--30 comparing the lm-eval harness baseline (dotted red line) against CORAL steering (green bars). Right: corresponding Expected Calibration Error (ECE) values. Middle layers (17--21) yield the strongest improvements in both metrics, while early layers (0--6) degrade performance below baseline. This pattern suggests correctness signals concentrate in layers where abstract semantic representations are most developed but output distributions remain malleable.
\vspace{-10pt}
\begin{figure}[hbt]  
    \centering
    \includegraphics[width=0.48\textwidth]{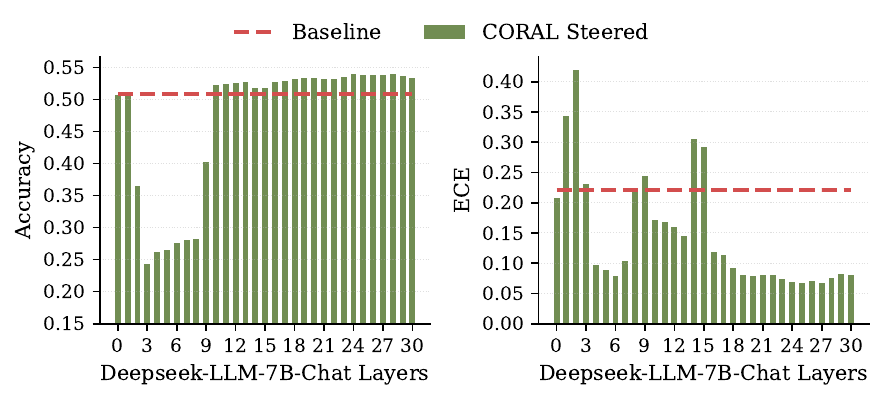}  
    \captionsetup{skip=-1.15pt}
    \caption{Layer-wise steering performance for DeepSeek-7B-Chat. Left: accuracy across layers 0--30 for \texttt{lm-eval} harness baseline (dotted red lines) and CORAL performance (green bars). Right: corresponding ECE values.}
    \label{fig:steer_layers}  
    \vspace{-22pt}
\end{figure}

Figure~\ref{fig:steer_layers} examines which layers contain the most useful correctness signals for steering. We train separate probes at each layer of DeepSeek-7B-Chat and evaluate steering performance independently.

% Middle layers (approximately 17-21) yield the strongest improvements in both accuracy and calibration, while early and late layers provide noisy outputs. At early layers (0-6), steering produces accuracy below the unsteered baseline. Performance improves steadily through layers 9-15, peaks in the middle layers, then marginally declines in the final layers. Calibration follows a similar trajectory. ECE drops sharply from early layers and remains consistently low through layers 19-24 before rising slightly again at the final layers.

Middle layers contain the most useful correctness signals for steering. Layers 17--21 yield the strongest improvements in both accuracy and calibration, while early and late layers provide noisy results. At early layers (0--6), steering produces accuracy below the unsteered baseline. Performance improves steadily through layers 9--15, peaks in the middle layers, then marginally declines in the final layers. Calibration follows a similar trajectory. ECE drops sharply from early layers and remains consistently low through layers 19--24 before rising slightly at the final layers.

This pattern aligns with prior work on representation formation in transformers \cite{sun2025transformerlayerspainters}. Early layers primarily handle syntactic processing and token-level features, while later layers commit to specific output tokens. The middle layers, where CORAL performs best, correspond to the stage where abstract semantic representations are most developed but output distributions are still somewhat malleable. 

\subsection{CORAL's Transferability}

A central claim of this work is that CORAL captures generalizable correctness directions rather than task-specific artifacts. Table~\ref{tab:transfer_results} evaluates this claim through transfer experiments on the complete published test sets of HellaSwag, OpenBookQA, ARC-Challenge, and Math-MC. 

CORAL demonstrates strong transferability across all four benchmark test sets and three model families, consistently exceeding baseline methods. On Mistral-7B-Instruct, CORAL achieves the highest accuracy on all transfer tasks while simultaneously improving calibration. The pattern holds for other models, where CORAL consistently achieves competitive or best performance across metrics.

Analyzing transfer patterns reveals that performance gains are strongest when source and target tasks share reasoning structure. Transfer to ARC-Challenge, which requires reasoning similar to CommonsenseQA, shows robust improvements: CORAL achieves 73.46\% accuracy on Mistral compared to 61.86\% for ITI and 63.14\% for CCPS. Transfer to Math-MC also performs well, suggesting that correctness signals generalize across domains requiring multi-step reasoning.

Transfer to HellaSwag presents a more nuanced picture. While CORAL achieves the best accuracy (81.98\% on Deepseek), the ECE improvements is only second best. This likely reflects a distributional shift: HellaSwag's answer options are substantially longer than those in probe training, making the probe's learned correctness direction less directly applicable.

Notably, CCPS shows limited transfer performance despite strong in-distribution results, underperforming few-shot and CORAL baselines on most transfer tasks. This contrast highlights that CORAL's regularized training objective successfully encourages learning of transferable representations rather than task-specific patterns.

% \begin{figure}[hbt]  
%     \centering
%     \includegraphics[width=0.5\textwidth]{images/reliability_diagram_comparison.pdf} 
%     \captionsetup{skip=0pt}
%     \caption{Reliability diagrams on \mm{insert} across three model families. The solid bars represent model baseline calibration and the dashed red stacked bar represents CORAL correction. The dashed gray line represents perfect calibration.}
%     \label{fig:reliability_diagram}  
% \end{figure}

%% file: mmlu_table.tex
\begin{table*}[!t]
\centering
\caption{Comparison of steering methods across different models and benchmarks. Blue rows are CORAL-steered results. \textbf{Bold} numbers indicate best performance and \underline{underlined} numbers indicate second best under Accuracy, ECE, cwECE, NLL, and Brier Score. Baseline methods are SteerConf \cite{zhou2025steerconf}, Few-Shot Prompting (Eval Harness) \cite{eval-harness}, Inference Time Intervention (ITI) \cite{li2024inference}, and CCPS \cite{10.1145/775047.775151, guo2017calibration, khanmohammadi-etal-2025-calibrating}.}
\label{tab:steering_results}
\setlength{\tabcolsep}{3pt}
\small

% ==================== MMLU Test Subset ====================
\textbf{MMLU Test Subset}\\[0.25em]
\resizebox{\textwidth}{!}{%
\begin{tabular}{@{}l ccccc@{}}
\toprule
\multicolumn{6}{c}{\textbf{Mistral-7B-Instruct-v0.3}} \\
\midrule
Method & Acc $\uparrow$ & ECE $\downarrow$ & cwECE $\downarrow$ & NLL $\downarrow$ & Brier $\downarrow$ \\
\midrule
SteerConf & 60.53 & 15.47 & \underline{7.16} & 146.54 & 20.66 \\
Fewshot & \underline{61.34} & 18.46 & 9.79 & 72.44 & 22.62 \\
ITI & 59.02 & 17.74 & 10.09 & 69.61 & 22.63 \\
CCPS & 61.20 & \underline{8.36} & 10.07 & \underline{55.49} & \underline{18.95} \\
\rowcolor{blue!5} CORAL & \textbf{64.22} & \textbf{3.27} & \textbf{2.83} & \textbf{54.30} & \textbf{17.92} \\
\bottomrule
\end{tabular}
\hspace{1em}
\begin{tabular}{@{}l ccccc@{}}
\toprule
\multicolumn{6}{c}{\textbf{Qwen2.5-7B-Instruct}} \\
\midrule
Method & Acc $\uparrow$ & ECE $\downarrow$ & cwECE $\downarrow$ & NLL $\downarrow$ & Brier $\downarrow$ \\
\midrule
SteerConf & 63.94 & 15.67 & \underline{8.75} & 113.96 & 20.57 \\
Fewshot & \underline{73.19} & 16.36 & 8.83 & 73.98 & 19.67 \\
ITI & 70.42 & 15.72 & 9.10 & 71.09 & 19.68 \\
CCPS & 73.02 & \underline{7.41} & 9.09 & \underline{56.67} & \underline{16.48} \\
\rowcolor{blue!5} CORAL & \textbf{74.23} & \textbf{4.23} & \textbf{2.77} & \textbf{46.61} & \textbf{15.51} \\
\bottomrule
\end{tabular}
\hspace{1em}
\begin{tabular}{@{}l ccccc@{}}
\toprule
\multicolumn{6}{c}{\textbf{Deepseek-7B-Chat}} \\
\midrule
Method & Acc $\uparrow$ & ECE $\downarrow$ & cwECE $\downarrow$ & NLL $\downarrow$ & Brier $\downarrow$ \\
\midrule
SteerConf & 37.48 & 23.39 & 13.89 & 91.15 & 30.46 \\
Fewshot & 50.84 & 22.12 & \underline{11.44} & 81.86 & 26.27 \\
ITI & 48.91 & 21.26 & 11.78 & 78.66 & 26.28 \\
CCPS & \underline{51.10} & \underline{8.38} & 12.05 & \underline{65.86} & \underline{22.60} \\
\rowcolor{blue!5} CORAL & \textbf{53.97} & \textbf{6.21} & \textbf{4.36} & \textbf{61.95} & \textbf{21.33} \\
\bottomrule
\end{tabular}%
}
\vspace{0.25em}

% ==================== RACE ====================
\textbf{RACE}\\[0.25em]
\resizebox{\textwidth}{!}{%
\begin{tabular}{@{}l ccccc@{}}
\toprule
\multicolumn{6}{c}{\textbf{Mistral-7B-Instruct-v0.3}} \\
\midrule
Method & Acc $\uparrow$ & ECE $\downarrow$ & cwECE $\downarrow$ & NLL $\downarrow$ & Brier $\downarrow$ \\
\midrule
SteerConf & 40.83 & 4.67 & 4.78 & 122.27 & 26.52 \\
Fewshot & 61.43 & 3.18 & 3.72 & \underline{57.59} & 19.72 \\
ITI & \underline{61.45} & 11.97 & 7.54 & 68.43 & \underline{17.77} \\
CCPS & 60.56 & \textbf{2.45} & \underline{3.29} & 57.80 & 19.77 \\
\rowcolor{blue!5} CORAL & \textbf{74.93} & \underline{3.08} & \textbf{3.18} & \textbf{55.06} & \textbf{15.63} \\
\bottomrule
\end{tabular}
\hspace{1em}
\begin{tabular}{@{}l ccccc@{}}
\toprule
\multicolumn{6}{c}{\textbf{Qwen2.5-7B-Instruct}} \\
\midrule
Method & Acc $\uparrow$ & ECE $\downarrow$ & cwECE $\downarrow$ & NLL $\downarrow$ & Brier $\downarrow$ \\
\midrule
SteerConf & 45.69 & 21.45 & 10.87 & 104.68 & 27.11 \\
Fewshot & 58.37 & 10.60 & \underline{6.44} & 64.84 & 21.48 \\
ITI & 58.34 & 15.60 & 13.06 & 70.45 & 19.36 \\
CCPS & \underline{58.38} & \underline{5.11} & \underline{6.44} & \underline{55.22} & \textbf{15.55} \\
\rowcolor{blue!5} CORAL & \textbf{71.04} & \textbf{3.46} & \textbf{3.35} & \textbf{54.40} & \underline{16.88} \\
\bottomrule
\end{tabular}
\hspace{1em}
\begin{tabular}{@{}l ccccc@{}}
\toprule
\multicolumn{6}{c}{\textbf{Deepseek-7B-Chat}} \\
\midrule
Method & Acc $\uparrow$ & ECE $\downarrow$ & cwECE $\downarrow$ & NLL $\downarrow$ & Brier $\downarrow$ \\
\midrule
SteerConf & 50.83 & 10.66 & 8.52 & 70.15 & 25.10 \\
Fewshot & \underline{55.67} & 4.40 & \underline{4.16} & 61.59 & 21.18 \\
ITI & 55.63 & 16.56 & 8.44 & 73.18 & \underline{19.09} \\
CCPS & 54.90 & \underline{3.00} & 4.25 & \underline{60.36} & 20.84 \\
\rowcolor{blue!5} CORAL & \textbf{73.06} & \textbf{2.62} & \textbf{3.08} & \textbf{57.69} & \textbf{16.06} \\
\bottomrule
\end{tabular}%
}
\vspace{0.25em}

% ==================== CommonsenseQA ====================
\textbf{CommonsenseQA}\\[0.25em]
\resizebox{\textwidth}{!}{%
\begin{tabular}{@{}l ccccc@{}}
\toprule
\multicolumn{6}{c}{\textbf{Mistral-7B-Instruct-v0.3}} \\
\midrule
Method & Acc $\uparrow$ & ECE $\downarrow$ & cwECE $\downarrow$ & NLL $\downarrow$ & Brier $\downarrow$ \\
\midrule
SteerConf & 29.24 & 33.71 & 14.09 & 142.01 & 34.56 \\
Fewshot & \underline{73.71} & 15.00 & \underline{6.74} & \underline{63.47} & 18.44 \\
ITI & 66.69 & 16.29 & 7.67 & 68.36 & 21.74 \\
CCPS & 72.98 & \underline{8.88} & 7.41 & 63.70 & \underline{18.32} \\
\rowcolor{blue!5} CORAL & \textbf{74.38} & \textbf{8.78} & \textbf{4.43} & \textbf{60.59} & \textbf{18.31} \\
\bottomrule
\end{tabular}
\hspace{1em}
\begin{tabular}{@{}l ccccc@{}}
\toprule
\multicolumn{6}{c}{\textbf{Qwen2.5-7B-Instruct}} \\
\midrule
Method & Acc $\uparrow$ & ECE $\downarrow$ & cwECE $\downarrow$ & NLL $\downarrow$ & Brier $\downarrow$ \\
\midrule
SteerConf & 39.81 & 18.54 & 7.77 & 134.80 & 20.16 \\
Fewshot & 84.68 & 12.16 & \underline{5.25} & 83.59 & 12.94 \\
ITI & 78.61 & 13.21 & 5.97 & 90.03 & 15.26 \\
CCPS & \underline{84.89} & \textbf{7.11} & \underline{5.25} & \underline{64.07} & \textbf{6.45} \\
\rowcolor{blue!5} CORAL & \textbf{85.01} & \underline{10.27} & \textbf{5.02} & \textbf{53.96} & \underline{9.75} \\
\bottomrule
\end{tabular}
\hspace{1em}
\begin{tabular}{@{}l ccccc@{}}
\toprule
\multicolumn{6}{c}{\textbf{Deepseek-7B-Chat}} \\
\midrule
Method & Acc $\uparrow$ & ECE $\downarrow$ & cwECE $\downarrow$ & NLL $\downarrow$ & Brier $\downarrow$ \\
\midrule
SteerConf & 38.57 & 13.12 & \underline{6.64} & 69.41 & 24.72 \\
Fewshot & 70.52 & 15.71 & 7.27 & 67.63 & \underline{19.61} \\
ITI & 63.80 & 17.06 & 8.27 & 72.84 & 23.12 \\
CCPS & \underline{70.53} & \underline{9.02} & 7.26 & \textbf{61.07} & 19.43 \\
\rowcolor{blue!5} CORAL & \textbf{72.17} & \textbf{8.54} & \textbf{4.62} & \underline{65.23} & \textbf{19.43} \\
\bottomrule
\end{tabular}%
}

\end{table*}

%% file: transfer_table.tex
\begin{table*}[!t]
\centering
\caption{Transfer performance of Probe 1 on four downstream benchmarks: HellaSwag, OpenBookQA, ARC-Challenge, and Math-MC. All three models show accuracy gains and calibration improvements across multiple independent benchmark tests. Blue rows are CORAL-steered results. \textbf{Bold} numbers indicate best performance and \underline{underlined} numbers indicate second best. Relevant other methods shown as comparison are SteerConf \cite{zhou2025steerconf}, Few-Shot Prompting (Eval Harness) \cite{eval-harness}, ITI \cite{li2024inference}, and CCPS (trained on CommonsenseQA) \cite{10.1145/775047.775151, guo2017calibration, khanmohammadi-etal-2025-calibrating}.}
\label{tab:transfer_results}
\setlength{\tabcolsep}{3pt}
\small

% ==================== Mistral-7B-Instruct-v0.3 ====================
\textbf{Mistral-7B-Instruct-v0.3 (Transfer Experiments})\\[0.25em]
\resizebox{\textwidth}{!}{%
\begin{tabular}{@{}l ccccc@{}}
\toprule
\multicolumn{6}{c}{\textbf{ARC-Challenge}} \\
\midrule
Method & Acc $\uparrow$ & ECE $\downarrow$ & cwECE $\downarrow$ & NLL $\downarrow$ & Brier $\downarrow$ \\
\midrule
SteerConf & 41.21 & 34.09 & 11.09 & 79.37 & 28.41 \\
Fewshot & \underline{61.86} & \underline{8.58} & 7.73 & \underline{62.52} & \underline{21.83} \\
ITI & \underline{61.86} & 26.01 & \underline{6.34} & 72.34 & 22.34 \\
CCPS & 63.14 & 24.60 & 7.40 & 85.34 & 28.83 \\
\rowcolor{blue!5} CORAL & \textbf{73.46} & \textbf{3.65} & \textbf{3.68} & \textbf{60.13} & \textbf{16.28} \\
\bottomrule
\end{tabular}
\hspace{0.5em}
\begin{tabular}{@{}l ccccc@{}}
\toprule
\multicolumn{6}{c}{\textbf{HellaSwag}} \\
\midrule
Method & Acc $\uparrow$ & ECE $\downarrow$ & cwECE $\downarrow$ & NLL $\downarrow$ & Brier $\downarrow$ \\
\midrule
SteerConf & 54.77 & 71.61 & 21.48 & 75.38 & 30.73 \\
Fewshot & \underline{81.23} & 32.88 & 15.89 & 67.88 & 24.34 \\
ITI & 74.43 & \underline{12.18} & \underline{8.62} & 63.43 & 21.81 \\
CCPS & 80.79 & 14.64 & 15.73 & \underline{59.44} & \underline{20.07} \\
\rowcolor{blue!5} CORAL & \textbf{82.35} & \textbf{3.10} & \textbf{2.23} & \textbf{45.95} & \textbf{14.60} \\
\bottomrule
\end{tabular}
\hspace{0.5em}
\begin{tabular}{@{}l ccccc@{}}
\toprule
\multicolumn{6}{c}{\textbf{OpenBookQA}} \\
\midrule
Method & Acc $\uparrow$ & ECE $\downarrow$ & cwECE $\downarrow$ & NLL $\downarrow$ & Brier $\downarrow$ \\
\midrule
SteerConf & 34.44 & 17.41 & 11.25 & 69.03 & 30.03 \\
    Fewshot & \underline{50.40} & \underline{11.40} & 8.80 & 70.18 & 24.50 \\
ITI & 50.23 & \underline{11.40} & \underline{7.91} & \textbf{60.50} & \underline{22.48} \\
CCPS & 52.00 & 28.92 & 8.10 & 92.09 & 32.97 \\
\rowcolor{blue!5} CORAL & \textbf{63.20} & \textbf{10.66} & \textbf{6.65} & \underline{69.57} & \textbf{20.70} \\
\bottomrule
\end{tabular}
\hspace{0.5em}
\begin{tabular}{@{}l ccccc@{}}
\toprule
\multicolumn{6}{c}{\textbf{Math-MC}} \\
\midrule
Method & Acc $\uparrow$ & ECE $\downarrow$ & cwECE $\downarrow$ & NLL $\downarrow$ & Brier $\downarrow$ \\
\midrule
SteerConf & 30.27 & 43.90 & 11.72 & 101.67 & 37.09 \\
Fewshot & \underline{44.89} & \underline{20.16} & 8.67 & \underline{91.55} & \underline{29.38} \\
ITI & 27.55 & 75.94 & 36.83 & 156.77 & 51.56 \\
CCPS & 44.40 & 34.57 & \textbf{2.45} & 105.13 & 37.51 \\
\rowcolor{blue!5} CORAL & \textbf{45.65} & \textbf{3.20} & \underline{2.78} & \textbf{65.42} & \textbf{23.12} \\
\bottomrule
\end{tabular}%
}
\vspace{0.25em}

% ==================== Qwen2.5-7B-Instruct ====================
\textbf{Qwen2.5-7B-Instruct - Transfer Experiments}\\[0.25em]
\resizebox{\textwidth}{!}{%
\begin{tabular}{@{}l ccccc@{}}
\toprule
\multicolumn{6}{c}{\textbf{ARC-Challenge}} \\
\midrule
Method & Acc $\uparrow$ & ECE $\downarrow$ & cwECE $\downarrow$ & NLL $\downarrow$ & Brier $\downarrow$ \\
\midrule
SteerConf & 42.80 & 19.12 & 8.93 & 77.89 & 26.92 \\
Fewshot & \underline{64.25} & \underline{4.84} & 6.22 & \underline{61.35} & \underline{20.69} \\
ITI & 63.93 & 14.67 & \underline{5.10} & 70.99 & 21.17 \\
CCPS & 62.37 & 8.55 & 6.49 & 65.21 & 22.98 \\
\rowcolor{blue!5} CORAL & \textbf{74.40} & \textbf{4.09} & \textbf{3.21} & \textbf{56.34} & \textbf{16.40} \\
\bottomrule
\end{tabular}
\hspace{0.5em}
\begin{tabular}{@{}l ccccc@{}}
\toprule
\multicolumn{6}{c}{\textbf{HellaSwag}} \\
\midrule
Method & Acc $\uparrow$ & ECE $\downarrow$ & cwECE $\downarrow$ & NLL $\downarrow$ & Brier $\downarrow$ \\
\midrule
SteerConf & 52.29 & 62.09 & 18.55 & 73.84 & 29.90 \\
Fewshot & \underline{77.56} & 28.51 & 13.72 & 66.49 & 23.68 \\
ITI & 71.07 & \underline{10.56} & 11.45 & 12.94 & 55.97 \\
CCPS & 76.61 & 11.45 & \underline{12.94} & \underline{55.97} & \underline{18.59} \\
\rowcolor{blue!5} CORAL & \textbf{81.41} & \textbf{10.43} & \textbf{9.59} & \textbf{49.80} & \textbf{16.06} \\
\bottomrule
\end{tabular}
\hspace{0.5em}
\begin{tabular}{@{}l ccccc@{}}
\toprule
\multicolumn{6}{c}{\textbf{OpenBookQA}} \\
\midrule
Method & Acc $\uparrow$ & ECE $\downarrow$ & cwECE $\downarrow$ & NLL $\downarrow$ & Brier $\downarrow$ \\
\midrule
SteerConf & 34.03 & 26.95 & 13.26 & 73.32 & 31.13 \\
Fewshot & \underline{49.80} & 17.64 & 10.37 & 74.54 & 25.40 \\
ITI & \underline{49.80} & \underline{17.64} & \underline{9.32} & \textbf{64.26} & \underline{23.31} \\
CCPS & 47.60 & 15.13 & 10.46 & 72.60 & 26.53 \\
\rowcolor{blue!5} CORAL & \textbf{65.40} & \textbf{7.21} & \textbf{7.44} & \underline{66.11} & \textbf{20.92} \\
\bottomrule
\end{tabular}
\hspace{0.5em}
\begin{tabular}{@{}l ccccc@{}}
\toprule
\multicolumn{6}{c}{\textbf{Math-MC}} \\
\midrule
Method & Acc $\uparrow$ & ECE $\downarrow$ & cwECE $\downarrow$ & NLL $\downarrow$ & Brier $\downarrow$ \\
\midrule
SteerConf & 38.28 & 11.15 & 4.31 & 71.62 & 27.89 \\
Fewshot & \underline{56.78} & \underline{5.12} & \underline{3.19} & \underline{64.49} & \underline{22.09} \\
ITI & 34.85 & 19.29 & 13.55 & 110.44 & 38.76 \\
CCPS & 55.74 & 22.58 & 3.71 & 80.57 & 29.32 \\
\rowcolor{blue!5} CORAL & \textbf{58.04} & \textbf{3.22} & \textbf{3.06} & \textbf{63.31} & \textbf{21.89} \\
\bottomrule
\end{tabular}%
}

% ==================== Deepseek-7B-Chat ====================

\vspace{0.25em}
\vspace{0.25em}

\textbf{Deepseek-7B-Chat - Transfer Experiments}\\[0.25em]
\resizebox{\textwidth}{!}{%
\begin{tabular}{@{}l ccccc@{}}
\toprule
\multicolumn{6}{c}{\textbf{ARC-Challenge}} \\
\midrule
Method & Acc $\uparrow$ & ECE $\downarrow$ & cwECE $\downarrow$ & NLL $\downarrow$ & Brier $\downarrow$ \\
\midrule
SteerConf & 46.09 & 24.64 & 14.72 & 86.28 & 30.81 \\
Fewshot & 54.18 & \underline{6.20} & 10.26 & \underline{67.96} & \underline{23.68} \\
ITI & 52.68 & 18.80 & \underline{8.41} & 78.64 & 24.23 \\
CCPS & \underline{54.35} & 15.13 & 10.41 & 74.82 & 27.08 \\
\rowcolor{blue!5} CORAL & \textbf{63.82} & \textbf{5.45} & \textbf{6.90} & \textbf{66.76} & \textbf{19.83} \\
\bottomrule
\end{tabular}
\hspace{0.5em}
\begin{tabular}{@{}l ccccc@{}}
\toprule
\multicolumn{6}{c}{\textbf{HellaSwag}} \\
\midrule
Method & Acc $\uparrow$ & ECE $\downarrow$ & cwECE $\downarrow$ & NLL $\downarrow$ & Brier $\downarrow$ \\
\midrule
SteerConf & 51.53 & 64.20 & 19.31 & 76.62 & 31.40 \\
Fewshot & \underline{76.43} & 29.48 & 14.28 & 68.99 & 24.87 \\
ITI & 70.03 & \textbf{10.92} & \underline{7.75} & 64.47 & \underline{22.28} \\
CCPS & 74.11 & 59.35 & 13.36 & 148.06 & 54.10 \\
\rowcolor{blue!5} CORAL & \textbf{81.98} & \underline{11.37} & \textbf{5.74} & \textbf{42.15} & \textbf{13.29} \\
\bottomrule
\end{tabular}
\hspace{0.5em}
\begin{tabular}{@{}l ccccc@{}}
\toprule
\multicolumn{6}{c}{\textbf{OpenBookQA}} \\
\midrule
Method & Acc $\uparrow$ & ECE $\downarrow$ & cwECE $\downarrow$ & NLL $\downarrow$ & Brier $\downarrow$ \\
\midrule
SteerConf & 32.80 & 26.15 & 14.76 & 83.71 & 30.48 \\
Fewshot & \underline{48.00} & \underline{17.12} & 11.55 & 85.11 & 24.87 \\
ITI & 46.48 & \underline{17.12} & \underline{10.38} & \textbf{73.37} & \textbf{22.82} \\
CCPS & 47.60 & 19.57 & 11.43 & \underline{80.47} & 28.73 \\
\rowcolor{blue!5} CORAL & \textbf{62.80} & \textbf{11.47} & \textbf{8.19} & 81.25 & \underline{23.19} \\
\bottomrule
\end{tabular}
\hspace{0.5em}
\begin{tabular}{@{}l ccccc@{}}
\toprule
\multicolumn{6}{c}{\textbf{Math-MC}} \\
\midrule
Method & Acc $\uparrow$ & ECE $\downarrow$ & cwECE $\downarrow$ & NLL $\downarrow$ & Brier $\downarrow$ \\
\midrule
SteerConf & 26.94 & 28.55 & 8.14 & 81.78 & 32.46 \\
Fewshot & \underline{39.95} & \underline{13.11} & 6.02 & \underline{73.64} & \underline{25.71} \\
ITI & 24.52 & 49.39 & 25.57 & 126.10 & 45.12 \\
CCPS & 39.93 & 40.93 & \underline{3.58} & 111.61 & 41.08 \\
\rowcolor{blue!5} CORAL & \textbf{43.73} & \textbf{6.46} & \textbf{3.37} & \textbf{66.21} & \textbf{23.47} \\
\bottomrule
\end{tabular}%
}
\end{table*}

%% file: sections/05_confidence_results.tex
\section{Sparse Autoencoders Analysis}
\label{sec:sae}

Having demonstrated CORAL's improvements, we investigate why a regularized probe captures correctness information effectively. We use Sparse Autoencoders (SAEs) to decompose activations into mono-semantic features and test whether isolated features can causally influence accuracy and calibration.

\subsection{SAE Methodology}

We train SAEs to decompose model activations into approximately mono-semantic features. The SAE encoder maps a z-score normalized activation $\mathbf{z} \in \mathbb{R}^{d}$ to a sparse code $\mathbf{f} \in \mathbb{R}^{D}$ where $D > d$:
\begin{equation}
\mathbf{f} = \text{ReLU}\left( \mathbf{W}_{\text{enc}} \mathbf{z} + \mathbf{b}_{\text{enc}} \right)
\end{equation}
The decoder reconstructs the activation from the sparse code:
\begin{equation}
\hat{\mathbf{z}} = \mathbf{W}_{\text{dec}} \mathbf{f} + \mathbf{b}_{\text{dec}}
\end{equation}
We train the SAE to minimize reconstruction error while encouraging sparsity through an L1 penalty weighted by decoder column norms:
\begin{equation}
\mathcal{L}_{\text{SAE}} = \|\mathbf{z} - \hat{\mathbf{z}}\|_2^2 + \lambda \sum_{j=1}^{D} |f_j| \cdot \|\mathbf{d}_j\|_2
\end{equation}
where $\mathbf{d}_j$ denotes the $j$-th column of $\mathbf{W}_{\text{dec}}$. We train SAEs at expansion ratios of $4\times$, $8\times$, and $16\times$ for a base dimension of $d = 4{,}096$. We provide detailed baseline implementation descriptions in Appendix~\ref{appendix:sae_detail}.

We evaluate selected features through causal ablation experiments. For each feature $j$, we remove its contribution from the reconstructed activation:
\begin{equation}
\mathbf{z}_{\text{ablated}} = \hat{\mathbf{z}} - f_j \mathbf{d}_j
\end{equation}
We use activation patching via forward hooks to replace the target layer's output with the ablated activation and measure the resulting change in accuracy and calibration. Features where ablation increases ECE are beneficial for calibration, while features where ablation decreases ECE are harmful.

\subsection{SAE Ablation Results}

\begin{figure}[hbt]  
    \centering
    \includegraphics[width=0.5\textwidth]{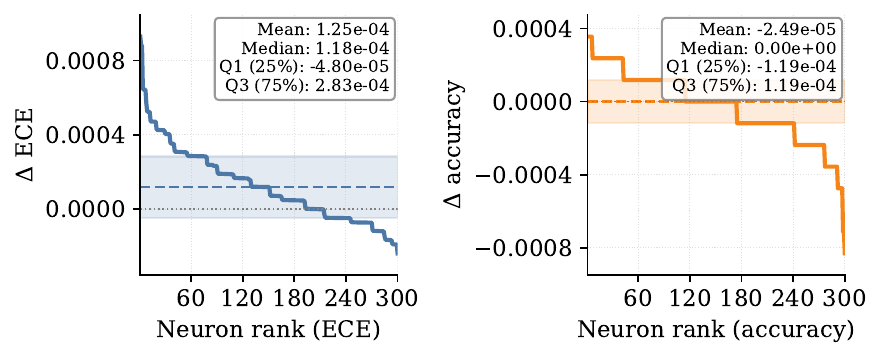}  
    \captionsetup{skip=-1pt}
    \caption{Distribution of single-neuron ablation impacts on ECE (left) and accuracy (right) for 300 SAE features selected by activation frequency and correlation with residual correctness. Shaded boxes show interquartile ranges and horizontal lines mark medians. Individual features produce negligible causal effects.}
    \label{fig:circuit_ablation}  
    \vspace{-1pt}
\end{figure}

Individual SAE features produce negligible causal effects on model calibration. Figure~\ref{fig:circuit_ablation} displays the distribution of ablation impacts across 300 neurons selected by correlation and activation-magnitude filtering. The mean ECE change from single-neuron ablation equals $1.25 \times 10^{-4}$, equivalent to 0.0125 percentage points, negligible compared to CORAL's 15+ percentage point improvements. The interquartile range spans $-4.80 \times 10^{-5}$ to $2.83 \times 10^{-4}$. Although we identify neurons that individually improve calibration when ablated (harmful features) and neurons that worsen calibration when ablated (beneficial features), their individual causal impacts remain negligible.

Aggregating beneficial features for steering also fails to recover meaningful gains. Steering with the top 50 to 300 features ranked by ablation impact does not improve accuracy or calibration and often degrades both metrics. For example, steering with the top 128 neurons yields a 0.12 percentage point decrease in accuracy and 0.14 percentage point increase in ECE. This pattern holds across feature counts and selection criteria.

Existing sparse steering methods show similarly limited effectiveness. Applying CorrSteer~\citep{cho2025corrsteer} to Deepseek-LLM-7B-Chat yields 51.0\% accuracy (matching baseline) and 22.01\% ECE (versus 22.12\% baseline) for MMLU test subset, a statistically insignificant improvement. The contrast with CORAL's substantial gains suggests that individual SAE features capture narrow semantic concepts correlated with correctness but do not encode a coherent correctness direction amenable to steering.

\section{Discussion}

\paragraph{Regularization Captures What Sparse Methods Miss.}
CORAL's performance stems from weight-decay regularization aggregating weak, distributed signals into coherent steering vectors. Our SAE ablations show that individual neurons produce negligible causal effects on MCQA correctness and aggregating top features by ablation impact fails to recover meaningful gains. This aligns with findings that correctness information exists as a collective property of the activation space \citep{lindsey2025biology}, requiring supervised aggregation rather than sparse feature selection.
\vspace{-1pt}

\paragraph{Transfer Results Suggest General Correctness Signals.}
Strong transfer across four complete independent benchmark test sets suggests that CORAL captures task-general correctness representations rather than dataset-specific patterns. Probes trained on CommonsenseQA and RACE transfer to the complete published test sets of ARC-Challenge, HellaSwag, OpenBookQA, and Math-MC without adaptation, spanning commonsense, science, abstraction, and mathematical reasoning. CCPS shows limited transfer despite strong in-distribution results, possibly because its contrastive training captures dataset-specific distinctions. CORAL's residual prediction objective may encourage learning features that generalize across reasoning types.

\vspace{-1pt}

\paragraph{Practical Efficiency for MCQA Deployment.}
CORAL trains in under 5 hours on a single GPU with 8.4k--10k labeled questions and adds minimal inference overhead. This efficiency contrasts with methods requiring model fine-tuning, multiple forward passes, or Jacobian computation. The concentration of correctness signals in middle layers (17--21) in our experiments suggests practitioners may reduce computation by targeting a single layer rather than scanning all layers.

%% file: sections/06_conclusion.tex
\section{Conclusion}

We present three key findings in this paper where we introduce CORAL: Correctness-Optimized Residual Activation Lens. First, regularized probes recover correctness signals from mid-layer activations reliably. MLP probes achieve accuracy gains of up to 20 points and ECE reductions of up to 14 points, with improvements generalizing across all three tested models. Second, learned steering directions transfer across tasks without retraining. On complete published benchmark test sets, the CORAL probe trained on CommonsenseQA and RACE activations averages 14\% accuracy improvements and 49\% ECE improvements. These improvements suggest CORAL identifies a general-purpose correctness subspace rather than task-specific decision boundaries. Third, correctness signals require supervised aggregation. Individual SAE features produce negligible causal effects and SAE-based steering degrades performance, while ridge and MLP probes successfully combine distributed signals into coherent steering vectors. CORAL requires only 5 hours of probe training on a single GPU and applies at inference time without modifying model weights.

% \section{Future Work}
% Several directions remain open for future work. Extending CORAL to open-ended generation tasks would test whether residual-correctness signals generalize beyond multiple-choice formats. Investigating the relationship between probe weights and attention patterns could reveal the computational mechanisms underlying correctness encoding. Combining CORAL with lightweight fine-tuning methods may yield further improvements beyond inference-time steering alone.

%% file: sections/07_limitations.tex
\section{Limitations}

CORAL's evaluation is limited to multiple-choice question answering, leaving open whether correctness signals generalize to free-form generation where correctness is harder to define. Additionally, CORAL requires labeled data to train probes, which may not be available for all domains or languages. Probe training is lightweight (minutes on a single GPU), but obtaining high-quality correctness labels at scale remains a practical constraint. At inference time, extracting activations and computing probe predictions adds latency. Finally, transfer performance depends on similarity between source and target tasks: transfer to tasks with substantially different reasoning patterns yields smaller improvements, suggesting the learned correctness direction is not fully task-agnostic.

\section{Impact Statement}

CORAL contributes to making large language model outputs more reliable. Improved calibration could benefit high-stakes domains such as medical diagnosis, legal analysis, and scientific research, where calibrated confidence enables systems to flag uncertain outputs for human review.

However, the same methodology could be adapted for harmful purposes. Malicious actors could train probes to steer models toward misinformation or suppress particular viewpoints, and the transferability we demonstrate means harmful steering vectors could generalize across tasks. We release our code and probe weights to enable reproducibility and encourage research into detection mechanisms for adversarial steering.

%% file: sections/appendix.tex
\section{Appendix}

\subsection{Sparse Autoencoder Implementation Details}
\label{appendix:sae_detail}

\paragraph{Feature Selection via Correlation Filtering.}
Inspired by CorrSteer \cite{cho2025corrsteer}, we identify calibration-relevant features by computing Pearson correlation between each SAE feature's activation and the residual correctness label. We first filter features by activation frequency, requiring a minimum of 10 active samples and activation frequency above 0.01\%. We define a feature as active when its magnitude exceeds $10^{-4}$. From the filtered set, we select the top-$k$ features with the highest positive correlation with the target variable. This correlation-based selection identifies features that linearly associate with calibration behavior. 

\paragraph{Feature Selection via Activation Magnitude.}
We implement an alternative selection method based on the feature's expected impact on downstream predictions. For each feature $j$, we compute its directional sensitivity as the alignment between its decoder column and the ridge probe weight vector:
\begin{equation}
s_j = \left( \mathbf{d}_j \odot \boldsymbol{\sigma} \right)^\top \mathbf{w}
\end{equation}
where $\odot$ denotes elementwise multiplication and $\boldsymbol{\sigma}$ is the activation standard deviation. We multiply this sensitivity by the mean feature activation to obtain an approximate impact score. We select features with the largest absolute impact scores, capturing directions that most influence the probe's residual prediction.

\paragraph{SAE Feature Steering.}

We implement SAE-based steering using forward hooks to modify hidden states during inference. We first identify beneficial features from the SAE ablation analysis results, where a feature is considered beneficial if ablating it decreases accuracy (negative accuracy impact) or increases ECE (positive ECE impact). For each beneficial feature $j$, we compute a steering weight:
\begin{equation}
w_j = \alpha_{\text{acc}} \cdot \max(-\Delta_{\text{acc},j}, 0) + \alpha_{\text{cal}} \cdot \max(\Delta_{\text{ECE},j}, 0)
\end{equation}
where $\Delta_{\text{acc},j} = \text{Acc}_{\text{ablated},j} - \text{Acc}_{\text{baseline}}$ is the change in accuracy when ablating feature $j$, and $\Delta_{\text{ECE},j} = \text{ECE}_{\text{ablated},j} - \text{ECE}_{\text{baseline}}$ is the corresponding change in calibration error. The hyperparameters $\alpha_{\text{acc}}$ and $\alpha_{\text{cal}}$ control the relative importance of accuracy and calibration contributions. We normalize these weights to sum to one across all selected features.

During inference, we register a forward hook at the target layer that applies additive steering. For each hidden state $\mathbf{h}$, we normalize using SAE training statistics, encode to obtain feature activations $f_j$, and compute the steering perturbation:
\begin{equation}
\mathbf{h}' = \mathbf{h} + \gamma \cdot \left( \sum_{j \in \mathcal{S}} f_j \cdot w_j \cdot \mathbf{d}_j \right) \odot \boldsymbol{\sigma}
\end{equation}
where $\mathcal{S}$ is the set of beneficial features, $\mathbf{d}_j$ is the decoder column for feature $j$, $\boldsymbol{\sigma}$ is the activation standard deviation, and $\gamma$ controls the global steering strength. This intervention amplifies features that causally improve calibration while preserving accuracy.

\subsection{Attention-Head Activations Analysis}
\label{sec:attention_head_analysis}
To investigate whether calibration-relevant information is localized to specific attention heads---as would be required for Inference-Time Intervention (ITI) style surgical steering ---we conducted a systematic probe analysis across all attention heads in the model.

\paragraph{Methodology.}
We extracted activations from the output of each attention head (before the output projection) at the last token position of each answer option. For DeepSeek-7B-Chat, this yields 960 total head representations (30 layers $\times$ 32 heads), each with dimension $d_{\text{head}} = 128$. Following the ITI approach, we trained separate 4-layer MLP probes on each (layer, head) pair to predict the residual correctness signal $r_i = y_i - \hat{p}_i$. Each probe uses hidden dimensions $(256, 128, 64, 32)$ with ReLU activations and dropout regularization. We evaluated generalization performance using 5-fold GroupKFold cross-validation (grouped by question to prevent leakage) and report $R^2$ scores on a held-out validation set.

\paragraph{Results.}
\begin{figure*}[t]  
    \centering
    \includegraphics[width=1.0\textwidth]{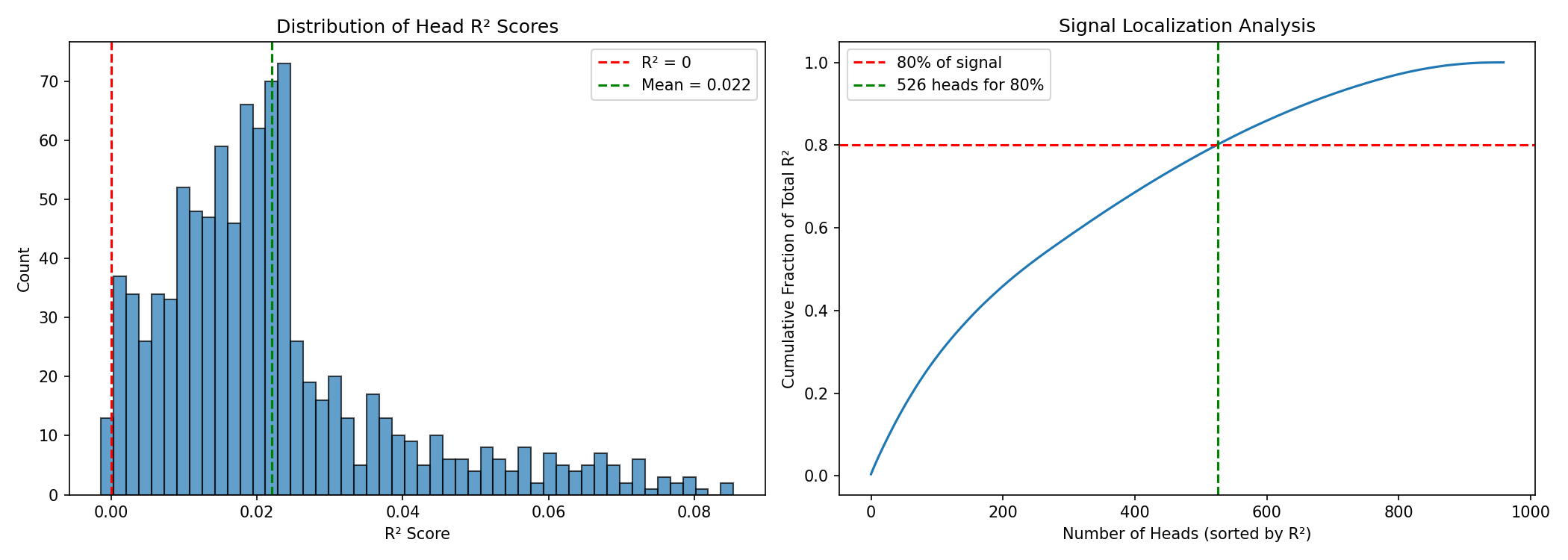}  
    \vspace{-5pt}
    \captionsetup{skip=2pt}
    \caption{Distribution of calibration signal across attention heads. \textit{Left:} Histogram of $R^2$ scores from 4-layer MLP probes trained on individual attention head activations to predict residual correctness. The mean $R^2 = 0.022$ and maximum $R^2 = 0.085$, with no head exceeding $R^2 > 0.10$. \textit{Right:} Cumulative signal analysis showing that 526 heads (55\% of all 960 heads) are required to capture 80\% of the total predictive signal, indicating that calibration information is distributed across the attention mechanism rather than localized to specific heads.}
    \label{fig:head_r2_distribution}  
    \vspace{-2pt}
\end{figure*}

\begin{figure}[t]
    \centering
    \includegraphics[width=\textwidth]{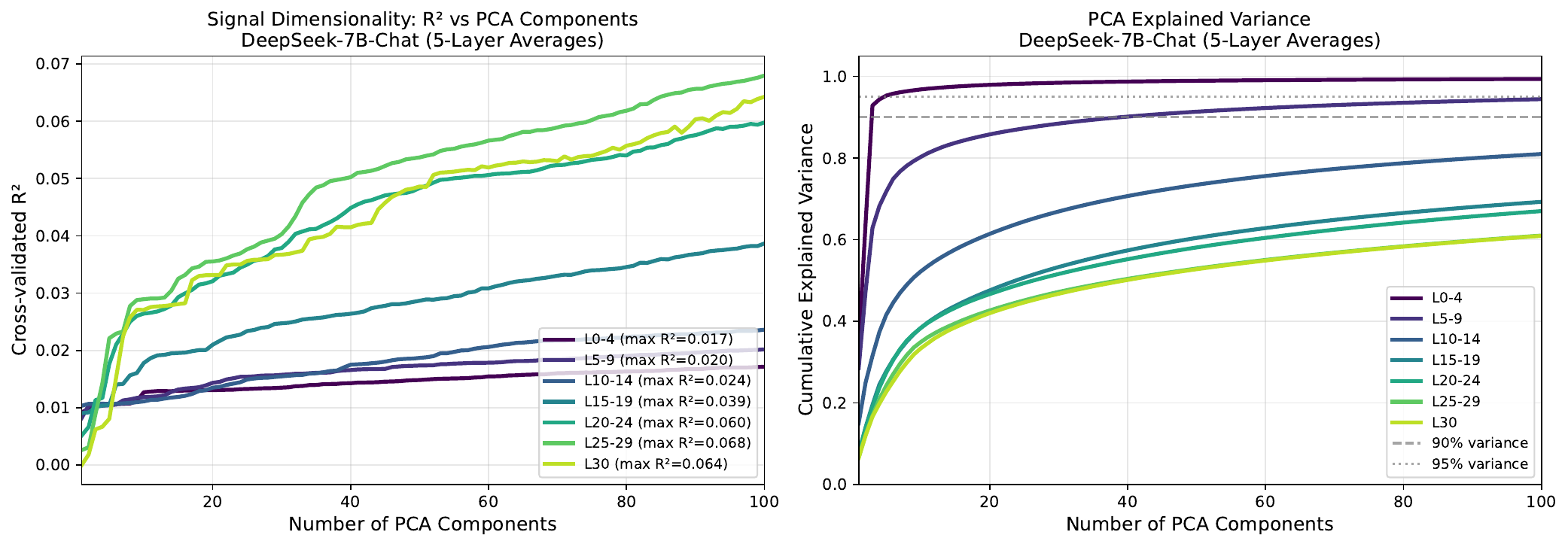}
    \caption{\textbf{Signal dimensionality analysis across layers.} \textit{Left:} Cross-validated $R^2$ for predicting residual correctness as a function of the number of PCA components, averaged over 5-layer groups. Later layers (L20--29) achieve higher predictive power, but all layer groups show gradual $R^2$ growth without saturation, indicating that the calibration signal is distributed across many dimensions rather than concentrated in a low-rank subspace. \textit{Right:} Cumulative explained variance of PCA components. Early layers exhibit highly concentrated activation variance (3 components capture $>$90\%), while later layers have more distributed representations. The mismatch between variance concentration and predictive power suggests calibration information resides in subtle activation patterns rather than dominant principal directions.}
    \label{fig:signal_dimensionality}
\end{figure}

Figure~\ref{fig:head_r2_distribution} shows the distribution of $R^2$ scores across all 960 attention heads. The left panel displays a histogram revealing that most heads have low predictive power, with a mean $R^2 = 0.022$ and maximum $R^2 = 0.085$. Notably, no individual head achieves $R^2 > 0.10$, and only 82 heads (8.5\%) exceed $R^2 > 0.05$.

The right panel presents a signal localization analysis: when heads are sorted by $R^2$ and their contributions are accumulated, we find that 526 heads (55\% of all heads) are required to capture 80\% of the total predictive signal. Equivalently, the top-48 heads---a reasonable target for surgical intervention---contain only 15.8\% of the total positive $R^2$. This indicates a highly distributed representation where calibration information is spread diffusely across the attention mechanism rather than concentrated in a small subset of ``calibration heads.''

\subsection{Signal Dimensionality Analysis}
\label{sec:signal_dimension_analysis}
To characterize the geometric structure of the calibration signal in activation space, we analyze how predictive power grows with the dimensionality of the representation. Specifically, we apply PCA to reduce the $d_{\text{model}} = 4096$ dimensional residual stream activations to $k$ principal components, then train Ridge regression probes to predict the residual correctness signal $r_i = y_i - \hat{p}_i$.

\paragraph{Methodology.}
For each layer, we fit PCA on the training activations and evaluate cross-validated $R^2$ as a function of the number of retained components $k \in \{1, 2, \ldots, 100\}$. If the calibration signal were low-dimensional (e.g., rank-1 or rank-10), we would expect $R^2$ to saturate quickly with few components. Conversely, a gradual increase in $R^2$ with $k$ indicates that the signal is distributed across many directions in activation space.

\paragraph{Results.}
Figure~\ref{fig:signal_dimensionality} shows the results averaged across groups of 5 consecutive layers. The left panel reveals that $R^2$ grows gradually with the number of PCA components across all layer groups, without saturating even at 100 components. Later layers (L20--24, L25--29) achieve higher maximum $R^2$ values (0.06--0.07), consistent with calibration information accumulating in deeper layers. The right panel shows cumulative explained variance: early layers (L0--4) have highly concentrated variance where 3 components capture over 90\% of activation variance, while later layers exhibit more distributed representations.

Importantly, despite the high concentration of \textit{activation variance} in few directions for early layers, the \textit{calibration signal} remains distributed---high explained variance does not imply high predictive power in those directions. This suggests that calibration-relevant information is encoded in subtle, distributed patterns rather than dominant principal components, motivating the use of full-dimensional MLP probes rather than low-rank approximations.

\section{Baseline Implementation Details}
\label{sec:baselines}

\subsection{Inference-Time Intervention (ITI) for Attention Head Steering}
\label{appendix:iti}

ITI~\citep{li2024inference} is an activation steering method that identifies and modifies specific attention head outputs during inference to improve model truthfulness.

\paragraph{Stage 1: Activation Collection.}
For each prompt $x$ in a binary-labeled dataset (e.g., TruthfulQA with correct/incorrect answer labels), we extract attention head activations at the last token position. Let $\mathbf{h}_{\ell,k} \in \mathbb{R}^{d_{\text{head}}}$ denote the output of attention head $k$ at layer $\ell$. We collect activations across all layers and heads using the output projection input:
\begin{equation}
    \mathbf{H} = \{\mathbf{h}_{\ell,k}^{(i)}\}_{\ell=1, k=1, i=1}^{L, K, N}
\end{equation}
where $L$ is the number of layers, $K$ is the number of heads per layer, and $N$ is the number of samples.

\paragraph{Stage 2: Probe Training and Head Selection.}
For each attention head $(\ell, k)$, we train a logistic regression probe to classify truthful vs.\ untruthful responses:
\begin{equation}
    p(y=1 \mid \mathbf{h}_{\ell,k}) = \sigma(\mathbf{w}_{\ell,k}^\top \mathbf{h}_{\ell,k} + b_{\ell,k})
\end{equation}
where $\mathbf{w}_{\ell,k} \in \mathbb{R}^{d_{\text{head}}}$ is the learned weight vector. The top-$M$ heads are selected based on validation accuracy, yielding $\mathcal{T} = \{(\ell_1, k_1), \ldots, (\ell_M, k_M)\}$.

\paragraph{Stage 3: Intervention Direction Computation.}
For each selected head, we compute the \emph{mass mean shift} direction as the difference between class centroids:
\begin{equation}
    \boldsymbol{\theta}_{\ell,k} = \frac{1}{|\mathcal{D}_+|}\sum_{i \in \mathcal{D}_+} \mathbf{h}_{\ell,k}^{(i)} - \frac{1}{|\mathcal{D}_-|}\sum_{i \in \mathcal{D}_-} \mathbf{h}_{\ell,k}^{(i)}
\end{equation}
where $\mathcal{D}_+$ and $\mathcal{D}_-$ denote the sets of truthful and untruthful samples, respectively. The direction is normalized: $\hat{\boldsymbol{\theta}}_{\ell,k} = \boldsymbol{\theta}_{\ell,k} / \|\boldsymbol{\theta}_{\ell,k}\|_2$.

\paragraph{Stage 4: Inference-Time Steering.}
During inference, we modify the attention head outputs at the last token position for all selected heads:
\begin{equation}
    \tilde{\mathbf{h}}_{\ell,k} = \mathbf{h}_{\ell,k} + \alpha \cdot \hat{\boldsymbol{\theta}}_{\ell,k}
\end{equation}
where $\alpha$ is a scalar hyperparameter controlling intervention strength. The intervention is applied using PyVene~\citep{wu-etal-2024-pyvene} by hooking into \texttt{model.layers[$\ell$].self\_attn.o\_proj.input}.

%%%%%%%%%%%%%%%%%%%%%%%%%%%%%%%%%%%%%%%%%%%%%%%%%%%%%%%%%%%%%%%
\subsection{Confidence-Conditioned Prediction with Shifting (CCPS)}
\label{appendix:ccps}

CCPS~\cite{khanmohammadi-etal-2025-calibrating} is a two-stage framework that learns to estimate response correctness by analyzing perturbation-based features extracted from model internals.

\paragraph{Stage 1: Perturbation-Based Feature Extraction.}
For each generated token $t$ at position $i$, we extract the hidden state $\mathbf{h}_i \in \mathbb{R}^{d}$ and logits $\mathbf{l}_i \in \mathbb{R}^{V}$ from the final layer before the language model head. We compute the Jacobian vector:
\begin{equation}
    \mathbf{J}_t = \nabla_{\mathbf{h}_i} \log p(t \mid \mathbf{h}_i)
\end{equation}

We generate a sequence of perturbed hidden states along the Jacobian direction:
\begin{equation}
    \tilde{\mathbf{h}}_i^{(\epsilon)} = \mathbf{h}_i + \epsilon \cdot \frac{\mathbf{J}_t}{\|\mathbf{J}_t\|_2}, \quad \epsilon \in \{\epsilon_1, \ldots, \epsilon_K\}
\end{equation}
and compute corresponding perturbed logits $\tilde{\mathbf{l}}_i^{(\epsilon)} = \text{LMHead}(\tilde{\mathbf{h}}_i^{(\epsilon)})$.

From these, we extract comprehensive features including:
\begin{itemize}[leftmargin=*,nosep]
    \item \textbf{Original state features}: log-probability, entropy, top-1/top-2 margin, hidden state norm
    \item \textbf{Perturbation sensitivity}: $\epsilon$-to-flip (minimum perturbation to change argmax), Jacobian norm
    \item \textbf{PEI (Perturbation Expected Information)}:
    \begin{equation}
        \text{PEI} = \frac{1}{K} \sum_{k=1}^{K} \max\left(0, \log p(t \mid \mathbf{h}_i) - \log p(t \mid \tilde{\mathbf{h}}_i^{(\epsilon_k)})\right)
    \end{equation}
    \item \textbf{Divergence metrics}: KL divergence, JS divergence, cosine similarity between original and perturbed distributions
\end{itemize}

\paragraph{Stage 2: Contrastive Embedding Learning.}
Let $\mathbf{f} \in \mathbb{R}^{d_f}$ denote the concatenated feature vector. We train an embedding network $g_\phi: \mathbb{R}^{d_f} \rightarrow \mathbb{R}^{d_e}$ using max-margin contrastive loss:
\begin{equation}
    \mathcal{L}_{\text{cont}} = \frac{1}{|\mathcal{B}|} \sum_{(i,j) \in \mathcal{B}} (1-y_{ij}) \cdot d_{ij}^2 + y_{ij} \cdot \max(0, m - d_{ij})^2
\end{equation}
where $d_{ij} = \|g_\phi(\mathbf{f}_i) - g_\phi(\mathbf{f}_j)\|_2$, $y_{ij} = \mathbbm{1}[y_i \neq y_j]$ indicates whether samples have different correctness labels, and $m$ is the margin hyperparameter.

The embedding network architecture is:
\begin{equation}
    g_\phi(\mathbf{f}) = \text{Linear}_{d_e}(\text{ReLU}(\text{Linear}_{64}(\text{ReLU}(\text{Linear}_{64}(\mathbf{f})))))
\end{equation}
with dropout ($p=0.1$) after each activation.

\paragraph{Stage 3: Correctness Classification.}
A classifier head is trained on top of the frozen or fine-tuned embeddings:
\begin{equation}
    p(y=1 \mid \mathbf{f}) = \text{softmax}(h_\psi(g_\phi(\mathbf{f})))
\end{equation}
where $h_\psi$ is an MLP with hidden dimensions [48, 24, 12]. Training uses cross-entropy loss with AdamW optimizer ($\text{lr}=10^{-4}$, weight decay $=0.01$).

\paragraph{Inference.}
For a new response, features are extracted from all generated tokens, standardized using the training scaler, and passed through the trained model. The probability $p(y=1 \mid \mathbf{f})$ serves as the confidence estimate.

%%%%%%%%%%%%%%%%%%%%%%%%%%%%%%%%%%%%%%%%%%%%%%%%%%%%%%%%%%%%%%%
\subsection{Confidence Steering via Prompt Shifting (SteerConf)}
\label{appendix:steerconf}

SteerConf~\cite{zhou2025steerconf} modulates LLM verbalized confidence through carefully designed prompt instructions that shift the model's risk tolerance.

\paragraph{Prompt Design.}
The base prompt template requests both an answer and a numerical confidence score:
\begin{quote}
\small
\texttt{Read the question, analyze step by step, provide your answer and your confidence in this answer. Note: The confidence indicates how likely you think your answer is true.\\
Use the following format to answer:\\
```Explanation: [insert step-by-step analysis here]\\
Answer and Confidence (0-100): [ONLY the \{answer\_type\}], [confidence]\%```}
\end{quote}

\paragraph{Confidence Shifting Instructions.}
We inject steering instructions that modulate the model's confidence calibration:
\begin{itemize}[leftmargin=*,nosep]
    \item \textbf{Very Cautious}: ``You are making important decisions, thus you should avoid giving a wrong answer with high confidence. You should be very cautious, and tend to give low confidence on almost all of the answers.''
    \item \textbf{Cautious}: ``You are making important decisions, thus you should avoid giving a wrong answer with high confidence.''
    \item \textbf{Confident}: ``You are making important decisions, thus you should avoid giving a right answer with low confidence.''
    \item \textbf{Very Confident}: ``You are making important decisions, thus you should avoid giving a right answer with low confidence. You should be very confident, and tend to give high confidence on almost all of the answers.''
\end{itemize}

\paragraph{Multi-Query Aggregation.}
For each question, we query the model $N$ times with temperature $\tau > 0$ to obtain answer-confidence pairs $\{(a_i, c_i)\}_{i=1}^N$. We compute aggregated metrics:

\textbf{Consistency Score}: The frequency of the most common answer:
\begin{equation}
    S_{\text{cons}} = \frac{|\{i : a_i = \hat{a}\}|}{N}, \quad \text{where } \hat{a} = \arg\max_a |\{i : a_i = a\}|
\end{equation}

\textbf{Weighted Confidence}: Combining mean confidence with consistency:
\begin{equation}
    S_{\text{weighted}} = \bar{c} \cdot (1 - V_{\text{ans}}) \cdot (1 - V_{\text{conf}})
\end{equation}
where $\bar{c} = \frac{1}{N}\sum_i c_i$ is the mean confidence, $V_{\text{ans}} = 1 - S_{\text{cons}}$ is answer variability, and $V_{\text{conf}} = \frac{\sigma_c / \bar{c}}{1 + \sigma_c / \bar{c}}$ is the normalized coefficient of variation of confidence scores.

\paragraph{Final Answer Selection.}
With majority voting, the final answer is selected as $\hat{a}$. If there are ties, we select the answer with the highest mean confidence among tied candidates.

%%%%%%%%%%%%%%%%%%%%%%%%%%%%%%%%%%%%%%%%%%%%%%%%%%%%%%%%%%%%%%%
\subsection{Hyperparameters and Training Details}
\label{appendix:hyperparams}

\paragraph{ITI.}
We select the top $M=48$ attention heads based on probe validation accuracy. The intervention strength $\alpha$ is tuned on a held-out validation set, typically in the range $[1.0, 3.0]$. Probes are trained with scikit-learn's \texttt{LogisticRegression} with \texttt{max\_iter=1000}.

\paragraph{CCPS.}
Contrastive model: hidden dimensions [64, 64], embedding dimension 8, ELU activation, dropout 0.05, margin $m=1.0$, trained for 5000 steps with batch size 64, learning rate $10^{-4}$. Classifier: hidden dimensions [48, 24, 12], trained for 5000 steps with learning rate $10^{-4}$. Feature extraction uses perturbation radii $\epsilon \in \{0.01, 0.1, 0.5, 1.0, 2.0, 5.0, 10.0, 20.0\}$.

\paragraph{SteerConf.}
Ensemble size $N=5$ with temperature $\tau=0.7$. We use chain-of-thought prompting for reasoning tasks. Confidence scores are normalized to $[0, 1]$ by dividing by 100.

%%%%%%%%%%%%%%%%%%%%%%%%%%%%%%%%%%%%%%%%%%%%%%%%%%%%%%%%%%%%%%%
\subsection{Baseline Tuning and Validation Protocols}
\label{appendix:baseline_tuning}

To ensure a fair comparison, all baseline methods were granted equivalent tuning flexibility to CORAL's per-model and per-dataset layer/$\gamma$ selection. We describe the tuning protocols for each method below.

\paragraph{ITI (Inference-Time Intervention).}
For each model and target dataset combination, we performed the following tuning:
\begin{itemize}[leftmargin=*,nosep]
    \item \textbf{Head selection}: Logistic regression probes were trained on a held-out validation split (20\% of training data) to select the top-$M$ attention heads. We tuned $M \in \{24, 48, 96\}$ per model.
    \item \textbf{Intervention strength}: The multiplier $\alpha$ was selected via grid search over $\{0.5, 1.0, 1.5, 2.0, 2.5, 3.0\}$ using validation ECE as the selection criterion.
    \item \textbf{Direction computation}: Both probe-weight directions and mass-mean-shift directions were evaluated; the better-performing variant was selected per model.
\end{itemize}
This mirrors the per-model tuning afforded to CORAL for layer selection and steering magnitude $\gamma$.

\paragraph{CCPS (Confidence-Conditioned Prediction with Shifting).}
CCPS was tuned with the following protocol:
\begin{itemize}[leftmargin=*,nosep]
    \item \textbf{Training data}: The contrastive embedding model and classifier were trained on a source dataset (e.g., CT-CHOICE) with an 80/20 train/validation split.
    \item \textbf{Architecture search}: Hidden dimensions for both the embedding network and classifier head were tuned per model using validation AUROC. We searched over embedding dimensions $\{8, 16, 32\}$ and classifier hidden layers from $\{[48, 24], [48, 24, 12], [64, 32, 16]\}$.
    \item \textbf{Per-dataset adaptation}: When transferring to a new target dataset, we allowed optional fine-tuning of the classifier head on a small validation subset (up to 500 samples) while keeping the contrastive encoder frozen.
    \item \textbf{Feature selection}: The perturbation radii $\{\epsilon_k\}$ and included feature types were validated per model to ensure stable feature extraction across different model architectures.
\end{itemize}

\paragraph{SteerConf (Confidence Steering via Prompt Shifting).}
SteerConf hyperparameters were tuned per model and dataset:
\begin{itemize}[leftmargin=*,nosep]
    \item \textbf{Prompt variant selection}: The steering prompt type (vanilla, cautious, very\_cautious, confident, very\_confident) was selected based on validation ECE for each model-dataset pair.
    \item \textbf{Sampling parameters}: Temperature $\tau \in \{0.5, 0.7, 1.0\}$ and ensemble size $N \in \{1, 3, 5, 10\}$ were tuned on a validation split.
    \item \textbf{Aggregation method}: The confidence aggregation strategy (mean, weighted, consistency-adjusted) was selected per dataset based on validation calibration metrics.
    \item \textbf{Chain-of-thought}: Whether to use CoT prompting was determined per dataset based on task complexity and validation performance.
\end{itemize}

\paragraph{Summary.}
All baseline methods were allowed comparable tuning flexibility to CORAL. Specifically, each method could select its key hyperparameters (ITI: heads and $\alpha$; CCPS: architecture and features; SteerConf: prompt type and sampling) independently for each model and dataset combination using held-out validation data. This ensures that performance differences reflect fundamental methodological advantages rather than tuning disparities.